\documentclass{article}

\usepackage[final]{neurips_2023}

\usepackage{framed,multirow}
\usepackage[utf8]{inputenc} 
\usepackage[T1]{fontenc}    
\usepackage{hyperref}       
\usepackage{url}            
\usepackage{booktabs}       
\usepackage{amsfonts}       
\usepackage{amsmath}
\usepackage{amssymb}
\usepackage{amsthm}         
\usepackage{nicefrac}       
\usepackage{microtype}      
\usepackage{xcolor}         
\usepackage{graphicx}
\usepackage{wrapfig}

\DeclareMathOperator*{\argmin}{arg\,min}

\title{Spatial-Temporal DAG Convolutional Networks for End-to-End Joint Effective Connectivity Learning \\and Resting-State fMRI Classification}

\author{%
  Rui Yang$^{1,2}$, Wenrui Dai$^1$, Huajun She$^1$, Yiping P. Du$^1$, Dapeng Wu$^2$, Hongkai Xiong$^1$ \\
  $^1$Shanghai Jiao Tong University, \; $^2$City University of Hong Kong\\
  \texttt{\{rui\_yang, daiwenrui, xionghongkai\}@sjtu.edu.cn}, \; \texttt{dapengwu@cityu.edu.hk} \\
}

\begin{document}

\maketitle

\begin{abstract}
Building comprehensive brain connectomes has proved of fundamental importance in resting-state fMRI (rs-fMRI) analysis. Based on the foundation of brain network, spatial-temporal-based graph convolutional networks have dramatically improved the performance of deep learning methods in rs-fMRI time series classification. However, existing works either pre-define the brain network as the correlation matrix derived from the raw time series or jointly learn the connectome and model parameters without any topology constraint. These methods could suffer from degraded classification performance caused by the deviation from the intrinsic brain connectivity and lack biological interpretability of demonstrating the causal structure (i.e., effective connectivity) among brain regions. Moreover, most existing methods for effective connectivity learning are unaware of the downstream classification task and cannot sufficiently exploit useful rs-fMRI label information. To address these issues in an end-to-end manner, we model the brain network as a directed acyclic graph (DAG) to discover direct causal connections between brain regions and propose Spatial-Temporal DAG Convolutional Network (ST-DAGCN) to jointly infer effective connectivity and classify rs-fMRI time series by learning brain representations based on nonlinear structural equation model. The optimization problem is formulated into a continuous program and solved with score-based learning method via gradient descent. We evaluate ST-DAGCN on two public rs-fMRI databases. Experiments show that ST-DAGCN outperforms existing models by evident margins in rs-fMRI classification and simultaneously learns meaningful edges of effective connectivity that help understand brain activity patterns and pathological mechanisms in brain disease.
\end{abstract}

\section{Introduction}
Resting-state functional magnetic resonance imaging (rs-fMRI), which detects the spontaneous fluctuations of Blood-Oxygen-Level-Dependent (BOLD) signals in the brain when a specific task is not being performed, has been extensively used to analyze the neural activity patterns of human brain~\cite{article1}. Over the past few decades, considerable effort has been devoted to estimating the intrinsic connectivity among brain regions of interest (ROIs) from spatial-temporal trajectories in rs-fMRI, which facilitates unveiling the operational principle of human brain architecture~\cite{article3,article2}. It has now been generally accepted that brain network analysis with the subject at rest plays a crucial part in studying brain development and diagnosis of brain diseases, such as attention deficit hyperactivity disorder~\cite{article70,article5}, Autism spectrum disorder~\cite{article6,article68} and major depressive disorder~\cite{article7,article71}.

Based on the constructed brain connectome, graph-based deep learning methods have emerged as effective tools to improve the classification accuracy of rs-fMRI time series, and are promising for disclosing the inherent working mechanism of human brain activity. For example, learning accurate predictors of a particular phenotype of the subjects (e.g., gender) has gained increasing attention in numerous rs-fMRI studies~\cite{article8,article9,article10,article11,article12,article38}. The core idea behind these studies is the spatial-temporal-based graph convolution, in which the structural information and temporal dynamics of brain graph signals are combined to learn expressive representation for the downstream gender prediction task. Despite achieving remarkable success, several problems still impede the effectiveness of spatial-temporal graph modeling in rs-fMRI analysis, as illustrated below.

Firstly, the brain graph in most existing spatial-temporal-based GCNs is determined based on the correlation (e.g., Pearson's correlation) between each pair of brain ROIs' BOLD signals~\cite{article8,article9,article10,article72}. Despite being widely adopted in modeling functional connectivity, the symmetric correlation matrix only reflects the simple linear relationships between brain ROIs and cannot explicitly reveal whether the connection is direct or indirect, which leads to a large gap between the estimated connectome and the real complex brain connectivity. Based on the previous studies claiming that graph convolutional networks (GCNs) strongly rely on the extracted graph topology for network analysis and their predictions are vulnerable to the perturbations of graph edges~\cite{article14}, we argue that the simple correlation matrix is inadequate for brain graph modeling and it is indispensable to learn a more complex and meaningful graph structure for more accurate rs-fMRI time series classification.

Secondly, albeit several recent studies~\cite{article7,article11,article38,article73} were dedicated to addressing the oversimplification problem of the correlation matrix by adaptively learning the graph adjacency matrices in spatial-temporal-based GCNs, they do not consider any explicit topology constraint. In essence, the learned graph structures in~\cite{article7,article11,article38,article73} and the correlation matrix are all complete graphs, thus lacking biological interpretability of reflecting the effective connectivity of human brain. Contrary to these functional connectivity networks that exhibit the simple fully-connected topology, brain effective connectivity is a causal graph comprising a smaller set of directed edges, in which edge weights characterize the direct causal influence that each neural area exerts over others~\cite{article37}. Inferring effective connectivity can be regarded as causal discovery from neuroimaging data such as rs-fMRI, which is one of the main end goals in brain network analysis and harder to solve than the simple correlation matrix~\cite{article2,article3,article37}. Moreover, effective connectivity networks of different population groups with and without certain disease commonly demonstrate distinct patterns and are important for assessing normal brain functions. A myriad of methods have centered on learning effective connectivity and provide promising perspectives for identifying the biomarkers related to certain brain disorder, e.g., Alzheimer's disease~\cite{article57,article35,article16,article36,article58,article39}. 

Most existing approaches only use the raw time series to infer the effective connectivity while additional rs-fMRI label information is not fully exploited. The small sample size, high noise level and low temporal resolution characteristics of fMRI time series data pose great challenges to the effectiveness and robustness of these methods~\cite{article37}. Recently, many studies~\cite{article8,article9,article10,article11,article12,article38} have demonstrated that the downstream gender prediction task can offer valuable supervised information from accurate rs-fMRI sex labels for training a GCN predictor to effectively discover and analyze human brain connectome, and more accurate and meaningful brain graph modeling could also benefit the performance of the GCN predictor. Based on this motivation, we aim to take advantage of both sides and propose a Spatial-Temporal DAG Convolutional Network (ST-DAGCN) to jointly learn the effective connectivity and enhance time series classification in rs-fMRI analysis.

To be concrete, the adjacency matrix of the directed connectome is treated as tunable parameters and is constrained to ensure the acyclicity, which is primarily dealt with in a broad range of effective connectivity modeling approaches~\cite{article63,article15,article57,article16,article37}. On such basis, we leverage the nonlinear structural equation model (SEM)~\cite{article65,article19} to design the spatial-temporal DAG convolution layer, which consists of a DAG convolution that captures the direct causal dependencies among different brain ROIs and an extra one-dimensional (1D) convolution that is cascaded to exploit temporal dynamics for each ROI's time series. Inspired by existing works on causal discovery~\cite{article17,article18,article19}, we formulate a continuous program for causal structure search and jointly optimize the DAG of effective connectivity and model parameters with a score-based learning method via gradient descent. The task-specific score combines the cross entropy between ST-DAGCN's predictions and rs-fMRI labels and an extra graph sparsity regularization. To demonstrate the advantages over existing spatial-temporal-based GCNs, we validate ST-DAGCN in the same gender classification task. ST-DAGCN is evaluated on two public rs-fMRI databases, i.e., disease-irrelevant Human Connectome Project (HCP) and disease-relevant AD (patients with Alzheimer's disease) and CN (control subjects) groups of Alzheimer's Disease Neuroimaging Initiative (ADNI). Experiments show that ST-DAGCN is effective in learning brain connectivity that boosts time series classification accuracy. Moreover, the extracted causal dependencies among brain ROIs indeed reveal some macroscopic structures as well as some local connectivity that echo previous neuroscience research for disclosing brain activity patterns and biomarkers of brain disease pathology. 

\section{Related Work}
\textbf{Spatial-Temporal-Based GCNs for Gender Prediction.}
Predicting a particular phenotype from a subject's rs-fMRI data has gained increasing interest in the neuroscience research community. A number of methods have been developed to solve the gender classification task based on the brain connectivity derived from rs-fMRI signals. Among them, graph-based deep learning has become the mainstream method that applies spatial-temporal-based GCNs to the constructed brain network and rs-fMRI time series to learn effective brain graph representation~\cite{article8,article9,article10,article11,article12,article38}. Two concurrent pioneering works~\cite{article8,article9,article10} firstly use the idea of Spatial Temporal Graph Convolutional Networks (ST-GCN)~\cite{article40} for rs-fMRI analysis, where the brain connectivity is built with the correlation matrix of brain ROIs' raw time series. To further improve the predictor's classification performance, STAGIN~\cite{article12} builds dynamic correlation matrices with sliding window approach and enhances the network's expressive power via spatial-temporal attention. DAST-GCN~\cite{article11} adaptively learns the graph adjacency matrix in each spatial-temporal graph convolution layer and uses dilated 1D convolution for temporal feature extraction. Since the learned adjacency matrices of brain graphs are different and essentially complete graphs at each layer, DAST-GCN does not have good interpretability in brain network analysis. FBNetGen~\cite{article38} employs a time series encoder followed by a graph generator to transform the time series into brain connectivity. For different subjects, it produces different connectome output, while our method learns a shared connectivity across a group of subjects, which enables us to easily identify the alteration in brain connectivity by comparing different populations (e.g., AD versus CN)~\cite{article57,article16,article69,article62}. Moreover, FBNetGen does not consider topology constraint and also learns fully-connected brain graphs like DAST-GCN, which impedes the network interpretability.

\textbf{Effective Connectivity Learning.}
Inferring effective connectivity from neuroimaging data is an ultimate goal of brain network analysis, as it offers a fresh perspective for disclosing the direct causation between brain regions. Most existing approaches can be divided into model-based and data-driven methods. Model-based methods require prior knowledge and depend on presupposed models to fit the neuroimaging data. Typical approaches include structural equation model (SEM)~\cite{article41}, dynamic causal modeling (DCM)~\cite{article42}, and multivariate autoregressive model (MVAR)~\cite{article43}. However, these methods are limited by model capacity and their performance is largely affected by hyperparameters. By comparison, data-driven methods directly infer effective connectivity from data without any prior knowledge or assumption. Among them, DAG is one common type of causal structure that is primarily dealt with~\cite{article15,article37}, and score-based learning, which uses specific score functions to measure the quality of candidate causal graphs and search for the one with the optimal score, has been widely used in DAG-based effective connectivity learning~\cite{article57,article16}. Recently, some deep learning approaches have been developed. nCREANN~\cite{article56} extends MVAR to the nonlinear case via artificial neural network. EC-GAN~\cite{article35} and EC-RGAN~\cite{article36} use GANs~\cite{article34} to learn effective connectivity. CVAEEC~\cite{article39} and STGCMEC~\cite{article58} utilize variational autoencoder~\cite{article44} and GCNs, respectively, to model the generative procedure of rs-fMRI time series. TBDS~\cite{article66} optimizes the GCN predictor and brain connectivity in an alternating manner, in which the downstream classification loss serves as the regularization term for maximum likelihood estimation of the raw time series in DAG-based effective connectivity learning. Therefore, TBDS does not directly learn task-specific connectivity in an end-to-end manner. Moreover, due to the introduction of the regularization term, the objective function may not satisfy the convergence condition of its employed optimization algorithm~\cite{article67}, causing the learned effective connectivity practically far from the DAG.

\section{Method}
\subsection{Preliminaries}
\textbf{Causal Discovery.}
In this paper, we adopt the score-based learning of DAG for inferring effective connectivity, which assigns a score to each candidate causal graph based on some pre-defined score function and outputs the DAG with the optimal score. Formally, we denote $\mathcal{G}=(\mathcal{V},\mathbf{A})$ as the causal graph, where $\mathcal{V}=\{v_i\}_{i=1}^{N}$ is the node set and $\mathbf{A}\in\mathbb{R}^{N\times N}$ is the weighted adjacency matrix. Each non-zero entry $\mathbf{A}[j,i]$ in $\mathbf{A}$ describes the strength of the direct causal influence $v_j \rightarrow v_i$. Given the pre-defined score function $F:\mathbb{R}^{N \times N}\rightarrow\mathbb{R}$, the causal search problem is defined as
\begin{equation}\label{eqn:score_discrete}
\min_{\mathbf{A}}\ F\left(\mathbf{A}\right) \quad\text{s.t.} \ \mathcal{G}\in\text{DAGs},
\end{equation}
which is a combinatorial optimization problem that is NP-hard in general and needs specialized search algorithm to solve over the combinatorial space of DAGs. To make the optimization procedure easier, the widely used strategy is to transform it into a continuous program~\cite{article17,article18,article19}:
\begin{equation}\label{eqn:score_continuous}
\min_{\mathbf{A}}\ F\left(\mathbf{A}\right) \quad \text{s.t.} \ h(\mathbf{A})=0,
\end{equation}
where $h:\mathbb{R}^{N \times N}\rightarrow\mathbb{R}$ is a smooth function for ensuring the acyclicity constraint with $h(\mathbf{A})=0$. By replacing the combinatorial constraint $\mathcal{G}\in\text{DAGs}$ with continuous constraint $h(\mathbf{A})=0$, the causal discovery problem has been transformed to a continuous optimization problem that can be efficiently solved via standard numerical algorithms such as augmented Lagrangian method. 

\textbf{Structural Equation Model (SEM).}
SEM is a popular model for describing the data generation procedure with causal relationships and is widely used in the estimation of effective connectivity~\cite{article41,article2,article3,article35,article36,article37,article39}. Suppose that, for the network system of $N$ nodes, each node $v_i$ is attached with a vector-valued signal or feature $\mathbf{s}_i\in\mathbb{R}^{d}$. The linear SEM takes the following form:
\begin{align} \label{sem}
\mathbf{s}_i = \sum_{j\in PA(i)} \mathbf{A}[j,i]\,\mathbf{s}_j + \epsilon_i,  
\end{align}
where $PA(i)$ denotes the parents of node $v_i$ induced from the causal DAG, i.e., signals are propagated from parent nodes to their children for aggregation, and $\epsilon_i\in\mathbb{R}^d (i=1,\ldots,N)$ are additive random noise (e.g., Gaussian noise) that are mutually independent between different nodes. 

\subsection{Problem Set-Up}
Given a subject's rs-fMRI data and an atlas, the brain is first parcellated into $N$ ROIs to extract their average BOLD time series $\mathbf{X}\in\mathbb{R}^{N\times T \times 1}$, where $T$ is the number of time points and each brain ROI's BOLD signal is standardized across time. Our goal is to learn a causal structure, i.e., brain effective connectivity, in gender classification for rs-fMRI. To be specific, we represent the brain network with a DAG $\mathcal{G}=(\mathcal{V},\mathbf{A})$, and here $\mathcal{V}=\{v_i\}_{i=1}^N$ is the set of brain ROIs and each $v_i$ is associated with a temporal BOLD signal $\mathbf{X}[i,:,:]$. Following~\cite{article8,article57,article16}, we assume the same brain (effective) connectivity is shared by all the subjects in the same group (e.g., all the subjects with AD). Based on the extracted time series $\mathbf{X}$ and the brain effective connectivity $\mathcal{G}$, ST-DAGCN is performed for gender prediction of the corresponding subject. Considering the fact that a better graph topology commonly leads to higher classification performance of GCNs~\cite{article14,neuralsparse,IDGL,emgcn}, we formulate the optimization problem as below, based on the causal discovery problem in~\eqref{eqn:score_continuous}, to jointly learn the causal structure $\mathbf{A}$ and the model parameters $\theta$ of ST-DAGCN.
\begin{equation}\label{eqn:opt}
\begin{aligned}
\min_{\mathbf{A},\theta} \quad & F\left(\mathbf{A},\theta\right) \equiv \frac{1}{M}\sum_{m=1}^{M}\mathcal{L}\left(y_{m}^\prime, y_{m}; \mathbf{A}, \theta\right) + \lambda\|\mathbf{A}\|_{1} \\
\text{s.t.} \quad & h(\mathbf{A}) \equiv \operatorname{tr}\left[(\mathbf{I}+\alpha \mathbf{A} \odot \mathbf{A})^{N}\right]-N=0.
\end{aligned}
\end{equation}
Here, $F\left(\mathbf{A},\theta\right)$ is the task-specific score function, in which $\mathcal{L}\left(y_{m}^\prime, y_{m}; \mathbf{A}, \theta\right)$ is the loss (i.e., cross entropy) between the ST-DAGCN's prediction $y^\prime$ and the target label $y$. $M$ is the number of training subjects. $\|\cdot\|_{1}$ denotes the $L_1$ norm for sparsity penalty on the causal graph. $\lambda$ is the balancing hyperparameter for sparsity-based regularization. $\operatorname{tr}\left[(\mathbf{I}+\alpha \mathbf{A} \odot \mathbf{A})^{N}\right]-N=0$ is the necessary and sufficient condition to ensure $\mathcal{G}\in\text{DAGs}$~\cite{article18}, in which $\odot$ denotes the Hadamard product of two matrices and $\alpha$ is a hyperparameter. Following~\cite{article18}, we define $\alpha=1/N$. 

\subsection{Spatial-Temporal DAG Convolutional Network}
The core of ST-DAGCN is the spatial-temporal DAG Convolution (ST-DAGC) layer for nonlinear structural equation modeling in rs-fMRI analysis. ST-DAGC consists of DAG convolution at the spatial level and 1D convolution at the temporal level such that the causal dependencies among brain ROIs and temporal dynamics of each ROI's time series are combined to learn effective representation for the downstream gender prediction task. In this section, we first introduce ST-DAGC and the network architecture for spatial-temporal graph modeling in rs-fMRI. Then we present the optimization procedure of ST-DAGCN to obtain the effective connectivity in gender prediction.

\newpage
\textbf{Spatial-Level Convolution.} 
At the spatial level, the linear SEM described by Eq.~\eqref{sem} is widely adopted in effective connectivity learning to model the causal dependencies between brain ROIs~\cite{article35,article36,article39}, but cannot capture the nonlinear relationships in feature generation. The linear SEM can be rewritten in the matrix form as $f(\mathbf{S},\mathbf{A}) = \mathbf{A}^\top \mathbf{S} + \mathbf{E}$, where $\mathbf{S}=[\mathbf{s}_1,\ldots,\mathbf{s}_N]^\top\in\mathbb{R}^{N\times d}$ is the feature matrix of $N$ ROIs and $\mathbf{E}=[\epsilon_1,\ldots,\epsilon_N]^\top\in\mathbb{R}^{N\times d}$ is the noise matrix. To increase the model expressivity, Ng et al.~\cite{article19} handle a more complex causal graph model that extends the linear SEM to the nonlinear case:
\begin{align} \label{nonlinear_sem}
f(\mathbf{S},\mathbf{A}) = g_2\left(\mathbf{A}^\top g_1\left(\mathbf{S}\right)\right),
\end{align}
where $g_2(\cdot)$ and $g_1(\cdot)$ are two neural networks for capturing nonlinear causal relationships. Considering that our model consists of multiple spatial-level DAG convolutions interleaved with nonlinear temporal-level convolutions, we use a linear transformation for $g_1(\cdot)$ and a nonlinear function for $g_2(\cdot)$ in Eq.~\eqref{nonlinear_sem} for one DAG convolution operation. Denote with $\mathbf{H}^{i}\in\mathbb{R}^{N\times T\times f}$ and $\mathbf{H}^{s}\in\mathbb{R}^{N\times T\times f_s}$ the input and output features of DAG convolution, where $f$ and $f_s$ are the numbers of input and output feature channels, and $\mathbf{H}^{i}=\mathbf{X}$ in the first ST-DAGC layer. For $t=1,\ldots,T$, we have
\begin{align} \label{eq1}
\mathbf{H}^{s}[:,t,:] &= f\left(\mathbf{H}^{i}[:,t,:],\mathbf{A}\right) \nonumber\\
&= \sigma\left(\operatorname{BatchNorm}\left(\mathbf{A}^\top\left(\mathbf{H}^{i}[:,t,:]\mathbf{W}^{s} + \mathbf{B}\right)\right)\right),
\end{align}
where $\mathbf{H}^{i}[:,t,:]\in\mathbb{R}^{N\times f}$ and $\mathbf{H}^{s}[:,t,:]\in\mathbb{R}^{N\times f_s}$ are the $t$-th slice of features before and after DAG convolution, $\mathbf{W}^{s}\in\mathbb{R}^{f\times f_s}$ is the learnable weights matrix, $\mathbf{B}\in\mathbb{R}^{N\times f_s}$ is the learnable bias matrix, $\operatorname{BatchNorm}\left(\cdot\right)$ is the batch normalization operation~\cite{article20} along the feature channel, and $\sigma\left(\cdot\right)$ is the nonlinear activation function. Under this feature generation rule in which each brain ROI's features are generated only dependent on causally relevant brain ROIs at each layer, DAG convolutions capture the nonlinear causal dependencies among brain ROIs by propagating nonlinear node features along the directed edges of effective connectivity for aggregation.

\textbf{Temporal-Level Convolution.}
1D convolution has proved to be an effective tool for analyzing the temporal dynamics of rs-fMRI data, as the BOLD signal itself is typically modeled as the 1D convolution of a stimulus function and the hemodynamic response function. A variety of sophisticated 1D convolution operations have been employed for rs-fMRI time series analysis, such as dilated convolution~\cite{article11} and causal convolution~\cite{article10}. In this work, we use the standard 1D convolution at the temporal level attributed to its efficiency and empirical remarkable performance. Specifically, subsequent to DAG convolution, the 1D convolution is cascaded to capture the temporal dynamics for each individual brain ROI. Denote $\mathbf{H}^t\in\mathbb{R}^{N\times T\times f_t}$ as the output features of 1D convolution and $f_t$ as the number of output feature channels. For $n=1,\ldots,N$:
\begin{align}
\mathbf{H}^{t}[n,:,:] = \sigma\left(\operatorname{BatchNorm}\left(\operatorname{Conv1D}\left(\mathbf{H}^{s}[n,:,:], \mathbf{W}^{t}\right)\right)\right),
\end{align}
where $\mathbf{H}^{s}[n,:,:]\in\mathbb{R}^{T\times f_s}$ and $\mathbf{H}^{t}[n,:,:]\in\mathbb{R}^{T\times f_t}$ are the $n$-th node features before and after 1D convolution, and $\mathbf{W}^{t}$ is the learnable 1D convolution filter. By analogy with the DAG convolution that aggregates information from the causally dependent spatial neighborhood for each brain ROI, 1D convolution actually also performs feature aggregation for each time point by considering the temporal neighborhood determined by the 1D convolution kernel. 

The above two-level convolution is the procedure for obtaining the output features $\mathbf{H}^i\rightarrow\mathbf{H}^s\rightarrow\mathbf{H}^t$ in an ST-DAGC layer, which serves as the building block for spatial-temporal nonlinear structural equation modeling in rs-fMRI analysis.

\textbf{Network Architecture of ST-DAGCN.}
Our designed ST-DAGCN architecture consists of three ST-DAGC layers to learn hierarchical graph representations with feature dimension set to $64$, i.e., $f=f_s=f_t=64$ except for $f=1$ in the first ST-DAGC layer. A global average pooling layer that computes the mean value within each feature channel is then performed on the resulting tensor to get a $64$-dimensional feature vector for each input rs-fMRI sequence. Finally, a fully-connected layer with a single output neuron activated by the sigmoid function is cascaded to generate the classification probability of gender prediction for the input subject's rs-fMRI data.

\textbf{Optimization.}
Inspired by score-based DAG learning approaches \cite{article17,article18,article19}, we use the augmented Lagrangian method to solve the causal structure learning problem in~\eqref{eqn:opt}. To be more specific, the augmented Lagrangian is defined as
\begin{align}\label{eq8}
L_{c}\left(\mathbf{A},\theta,\eta\right) = F\left(\mathbf{A},\theta\right) + \eta h(\mathbf{A}) + \frac{c}{2}|h(\mathbf{A})|^2,
\end{align}
where $\theta$ is the set of all the learnable parameters of ST-DAGCN, $\eta$ is the Lagrange multiplier, and $c>0$ is the penalty parameter. At the $k$-th iteration of optimization, three steps are performed to alternately update $\{\mathbf{A}^{(k)},\theta^{(k)}\}$, $\eta^{(k)}$, and $c^{(k)}$:
\begin{align}
\{\mathbf{A}^{(k)},\theta^{(k)}\} &= \argmin_{\mathbf{A},\theta} L_{c^{(k)}}\left(\mathbf{A},\theta,\eta^{(k)}\right), \label{eq4}\\
\eta^{(k+1)} &= \eta^{(k)} + c^{(k)}h(\mathbf{A}^{(k)}), \label{eq5}\\
c^{(k+1)} &= \left\{
\begin{aligned}\label{eq6}
& \beta c^{(k)} , &\text{if} \ |h(\mathbf{A}^{(k)}|>\gamma|h(\mathbf{A}^{(k-1)}|, \\
& c^{(k)},  &\text{otherwise},
\end{aligned}
\right.
\end{align}
where $\beta>1$ and $\gamma<1$ are hyperparameters. The subproblem~\eqref{eq4} can be solved via gradient descent. Specifically, we use Adam optimizer~\cite{article22} implemented by PyTorch. To avoid over-fitting in solving~\eqref{eq4}, dropout~\cite{article21} is employed after each ST-DAGC layer, and we perform $L_2$ regularization on all the weights matrices. The optimization procedure guarantees that $\mathbf{A},\theta$, and $\eta$ can converge to local optimum, and $c$ progressively increases such that $h(\mathbf{A})$ vanishes to make $\mathbf{A}$ approximate a DAG. Thus, the final output DAG is the causal graph that achieves the optimal score in helping ST-DAGCN classify rs-fMRI time series. We provide detailed implementations in the Appendix.

\textbf{Post-Processing.}
Since $h(\mathbf{A})$ cannot be strictly zero at the end of optimization, post-processing is necessary to produce the exact DAG that reveals the potential causal dependencies among different brain ROIs. To be concrete, we first threshold the learned brain connectivity by $\mathbf{A}\odot1[|\mathbf{A}|>\varepsilon]$ to remove probably falsely discovered connections, where $1[\cdot]$ is the element-wise Iverson bracket and $\varepsilon>0$ is the threshold value. If the thresholded adjacency matrix still contains cycles, we use depth-first search (DFS) to find the cycle and remove the edge with the smallest absolute value in the cycle, and repeat this procedure until no cycles can be found in the final processed graph $\mathbf{A}_{\text{DAG}}$. The resulting residual caused by $\mathbf{A}-\mathbf{A}_{\text{DAG}}$ can be viewed as the noise matrix part in SEM.

\section{Experiments and Analysis}
\subsection{Datasets and Baselines}
In this study, we evaluate all the methods on two publicly available rs-fMRI databases, i.e., Human Connectome Project (HCP)~\cite{article23} and Alzheimer’s Disease Neuroimaging Initiative (adni.loni.usc.edu). The evaluation metrics and scheme are provided in the Appendix.

\textbf{HCP}. For a fair comparison with the previous spatial-temporal graph modeling approach like ST-GCN~\cite{article8}, we use its public pre-processed data that consists of 1089 rs-fMRI scans (496 males and 593 females) with repetition time $\text{TR}=0.72\text{s}$ and the total number of time points $T_{\text{total}}=1000$. The brain is parcellated into $N=22$ ROIs according to~\cite{article25}, and the average BOLD signal in each ROI is standardized to z-scores. 

\textbf{ADNI}. We select 132 subjects that are divided into two groups: CN (control subjects) and AD (patients with Alzheimer’s disease). The CN group contains 32 females and 40 males, and the AD group consists of 31 females and 29 males. The brain is parcellated into $N=39$ ROIs via the MSDL atlas~\cite{article26}. The rs-fMRI data is acquired with the same scanner of $\text{TR}=3\text{s}$. Each brain ROI's BOLD signal is also standardized to z-scores and $T_{\text{total}}=130$.

\textbf{Baselines.} We compare with various spatial-temporal modeling approaches for rs-fMRI: (1) The LSTM model~\cite{article27} trained on the BOLD time series. (2) ST-GCN~\cite{article8,article9,article10} that uses 1D convolution at the temporal level and graph convolution at the spatial level based on the pre-defined Pearson's correlation matrix. The specific implementation of convolution is selected from~\cite{article8,article9,article10} and tuned for each dataset to achieve the best accuracy. (3) DAST-GCN~\cite{article11} that adaptively learns an adjacency matrix in each spatial graph convolution layer and uses dilated 1D convolution at the temporal level. (4) STAGIN~\cite{article12} that calculates dynamic Pearson's correlation and attentive graph read-out at the spatial level and employs Transformer-based attention~\cite{article28} at the temporal level. The original STAGIN model uses the Pearson's correlation matrices in which only the top 30-percentile values are kept. We also consider the variant using the full correlation (FC) matrices. (5) FBNetGen~\cite{article38} in which a time series encoder is employed to learn temporal features and a graph generator is followed to build the brain network based on the encoded features. The correlation matrix and the learned brain network will serve as the input node features and adjacency matrix for a GCN predictor.

\subsection{Results on Disease-Irrelevant HCP Dataset}
\textbf{Gender Prediction.}
Table~\ref{tab1} reports the classification results on the HCP dataset. Overall, we make the following observations. (1) Our ST-DAGCN achieves the best classification performance on the HCP database. (2) With the DAG topology constraint based on nonlinear SEM, ST-DAGCN significantly outperforms existing graph-based spatial-temporal modeling approaches. Noticeably, ST-DAGCN achieves about 5\% gain in mean ACC and 4\% gain in mean AUC compared with the runner-up ST-GCN without topology constraint. (3) Both the correlation matrix and previous adaptive graph learning manners cannot guarantee that the spatial-temporal-based GCNs reach their greatest potential in rs-fMRI gender prediction. (4) In experiments, we find that the complex and large-scale STAGIN models tend to suffer from over-fitting and perform badly on HCP pre-processed by~\cite{article8}. In contrast, our model demonstrates better generalization ability in terms of four evaluation metrics.

\begin{table}[t]
\vspace{-4pt}
\setlength{\tabcolsep}{12pt}
\centering
\scriptsize
\setlength{\belowcaptionskip}{1pt}
\caption{rs-fMRI time series classification results on the HCP dataset.}\label{tab1}
\begin{tabular}{lcccc}
\toprule
Method & ACC (\%) & SEN (\%) & SPE (\%) & AUC (\%)  \\
\midrule
LSTM & 71.53 $\pm$ 1.80  & 60.88 $\pm$ 5.75  & 80.44 $\pm$ 3.14   & 74.98 $\pm$ 2.51 \\
ST-GCN & 79.52 $\pm$ 2.33 & 76.20 $\pm$ 7.28  & 82.31 $\pm$ 6.70  & 85.99 $\pm$ 1.70 \\
DAST-GCN & 78.24 $\pm$ 2.99  & 71.97 $\pm$ 6.39  & 83.47 $\pm$ 4.05  & 82.43 $\pm$ 3.85 \\
STAGIN (Original) & 78.14 $\pm$ 1.95  & 69.74 $\pm$ 8.45  & 85.13 $\pm$ 6.57  & 84.29 $\pm$ 2.85 \\
STAGIN (FC) & 76.58 $\pm$ 2.12  & 72.17 $\pm$ 7.72  & 80.27 $\pm$ 2.81  & 81.17 $\pm$ 3.02 \\
FBNetGen & 66.67 $\pm$ 0.83  & 57.85 $\pm$ 8.73  & 74.04 $\pm$ 8.09   & 67.16 $\pm$ 1.16 \\
ST-DAGCN & 85.03 $\pm$ 2.25  & 84.87 $\pm$ 2.16  & 85.16 $\pm$ 3.64   & 89.90 $\pm$ 2.33 \\
\bottomrule
\end{tabular}
\end{table}
\begin{figure}[t]
\vspace{-6pt}
\centerline{\includegraphics[scale=0.3]{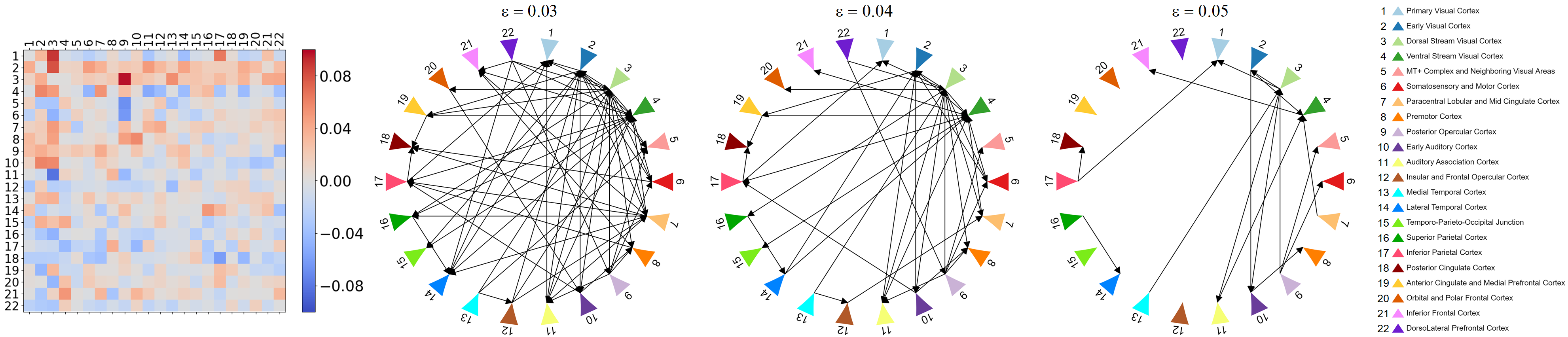}}
\setlength{\abovecaptionskip}{-1pt}
\caption{Learned effective connectivity on the HCP dataset, which consists of 22 brain ROIs~\cite{article25} pre-processed by~\cite{article8}. The left-most figure is the averaged $\mathbf{A}^\top$, and the right three figures are the produced DAGs with different threshold values.}
\label{hcp_dag}
\vspace{-2pt}
\end{figure}

\textbf{Effective Connectivity Analysis.} Here, we visualize and analyze the final learned effective connectivity. We optimize the causal structure learning problem with all the subjects' data for 10 trials and then average the learned $\mathbf{A}$ for producing the DAG with post-processing. Figure~\ref{hcp_dag} visualizes the learned effective connectivity. We find that the visual regions (nodes 1-5) have more neural connections and this result is consistent with previous studies~\cite{article30,article29} claiming that the visual cortex is highly correlated with rs-fMRI information pattern. The dorsal stream visual cortex (node 3) involves the most direct links with other neural areas, which is aligned with its functionality that it includes higher visual regions responsible for visual stimuli perception and plan of visually guided actions~\cite{article25}. Moreover, our discoveries are also in agreement with previous research~\cite{article8,article33} showing that visual regions are crucial in demonstrating gender-related differences in rs-fMRI. These findings physiologically verify that the produced effective connectivity not only substantially improves the rs-fMRI classification performance, but also learns some meaningful macroscopic structures in reflecting the causal dependencies among brain ROIs to help understand normal brain functions.

\subsection{Results on Disease-Relevant ADNI Dataset}
Existing works separately infer the effective connectivity of AD and CN groups for the downstream analysis ~\cite{article36,article16,article39}. Following this paradigm, we apply ST-DAGCN to the AD and CN groups of ADNI database, respectively, to learn the effective connectivity and perform gender classification.

\textbf{Gender Prediction.} We provide the gender classification results in Table~\ref{tab2}. As can be seen, our proposed ST-DAGCN consistently performs best in terms of the ACC and AUC metrics on both AD and CN groups. We can also observe that ST-DAGCN outperforms existing graph modeling approaches without topology constraint by large margins. Averagely, ST-DAGCN achieves over 3.0\% and 5.5\% accuracy improvement against previous state-of-the-art methods on the AD and CN groups respectively. This result also verifies that more effective modeling of brain connectivity is of great benefit to the spatial-temporal-based GCNs in rs-fMRI time series classification.

\begin{table}[t]
\vspace{-4pt}
\setlength{\tabcolsep}{3pt}
\centering
\tiny
\setlength{\belowcaptionskip}{1pt}
\caption{rs-fMRI time series classification results on the AD and CN groups of the ADNI dataset.}\label{tab2}
\begin{tabular}{lcccccccc}
\toprule
\multirow{2}{*}{Method} & \multicolumn{4}{c}{AD} & \multicolumn{4}{c}{CN} \\
\cmidrule(lr){2-5} \cmidrule(lr){6-9}
& ACC (\%) & SEN (\%) & SPE (\%) & AUC (\%)  & ACC (\%) & SEN (\%) & SPE (\%) & AUC (\%) \\
\midrule
LSTM & 81.67 $\pm$ 8.16  & 76.00 $\pm$ 7.72  & 87.14 $\pm$ 12.38  & 79.33 $\pm$ 12.85 & 75.14 $\pm$ 10.9   & 77.50 $\pm$ 21.51   & 70.95 $\pm$ 17.13   & 71.96 $\pm$ 15.16 \\
ST-GCN & 81.67 $\pm$ 3.33   & 83.33 $\pm$ 14.91   & 80.95 $\pm$ 11.57   & 73.17 $\pm$ 5.48  & 79.14 $\pm$ 6.22   & 82.50 $\pm$ 12.75   & 75.24 $\pm$ 15.47  &  78.10 $\pm$ 11.94 \\
DAST-GCN & 80.00 $\pm$ 8.50   & 76.00 $\pm$ 7.72   & 83.81 $\pm$ 10.58   & 78.22 $\pm$ 9.34  & 77.62 $\pm$ 5.75   & 85.00 $\pm$ 12.25   & 68.10 $\pm$ 11.43   & 75.83 $\pm$ 9.98  \\
STAGIN (Original) & 80.00 $\pm$ 4.08  & 80.00 $\pm$ 16.33  & 80.95 $\pm$ 18.87  & 79.32 $\pm$ 4.22  & 76.29 $\pm$ 5.81 & 75.00 $\pm$ 17.68   & 77.14 $\pm$ 17.21  & 72.02 $\pm$ 6.54 \\
STAGIN (FC) & 81.67 $\pm$ 3.33  & 83.33 $\pm$ 10.54  &  80.95 $\pm$ 11.57  & 79.35 $\pm$ 7.11  & 80.48 $\pm$ 3.14  & 87.50 $\pm$ 11.18    & 70.95 $\pm$ 19.37   & 75.24 $\pm$ 9.32 \\
FBNetGEN & 70.00 $\pm$ 4.08  & 70.00 $\pm$ 22.11  & 71.43 $\pm$ 14.75  & 66.81 $\pm$ 6.03 & 73.71 $\pm$ 4.52   & 75.00 $\pm$ 13.69   & 72.38 $\pm$ 14.57   & 75.33 $\pm$ 4.55 \\
ST-DAGCN  & 85.00 $\pm$ 6.24  & 89.33 $\pm$ 8.79    & 80.48 $\pm$ 16.45   & 79.87 $\pm$ 9.88 & 86.10 $\pm$ 4.54   & 92.50 $\pm$ 6.12   & 77.62 $\pm$ 13.85   & 83.21 $\pm$ 7.53 \\
\bottomrule
\end{tabular}
\end{table}
\begin{figure}[t]
\vspace{-6pt}
\centerline{\includegraphics[scale=0.29]{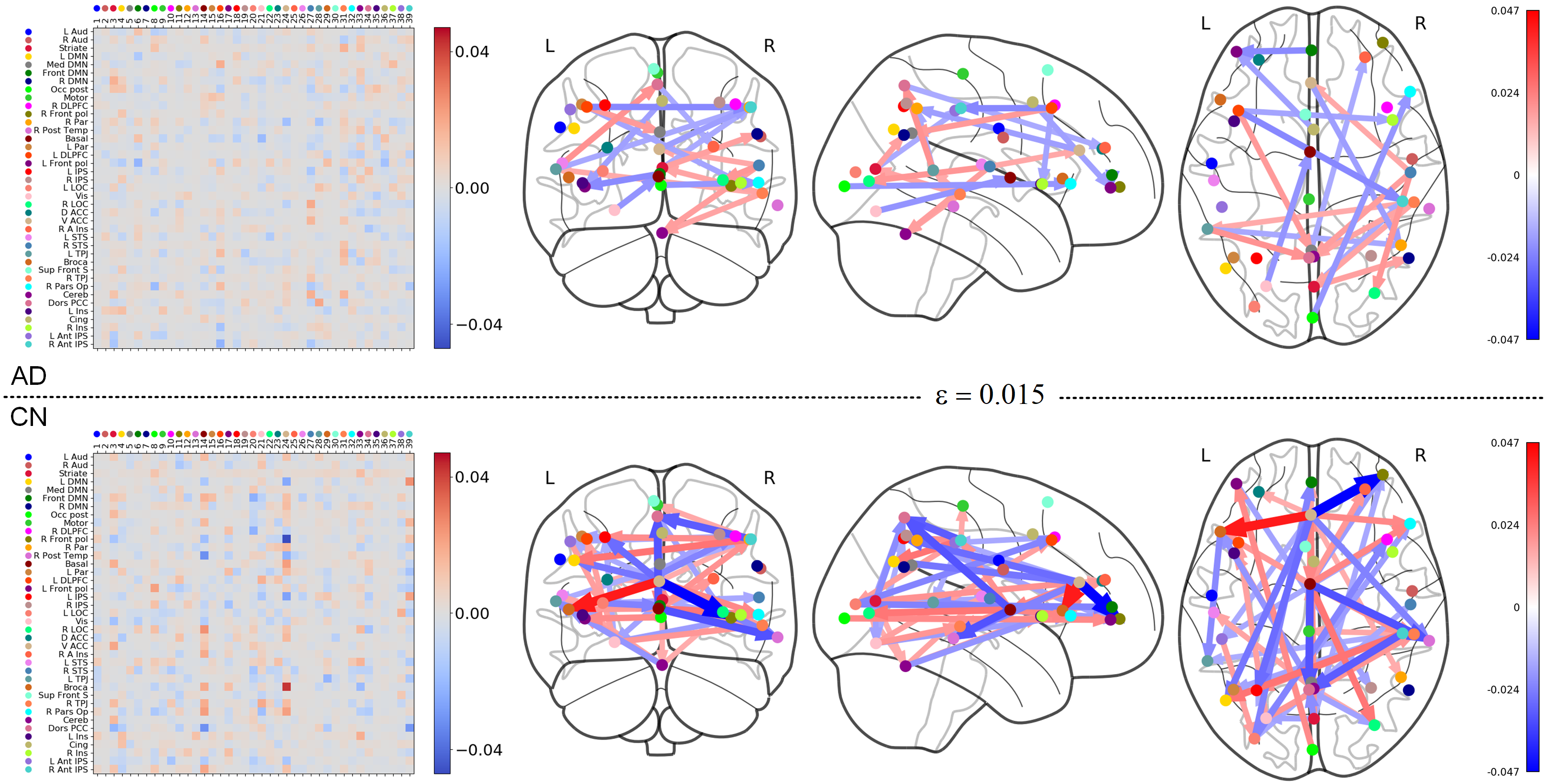}}
\setlength{\abovecaptionskip}{-1pt}
\caption{Estimated effective connectivity on the ADNI dataset. The brain is parcellated into 39 ROIs according to the MSDL atlas~\cite{article26}. The left two matrices are the averaged $\mathbf{A}^\top$, and the right three columns are estimated brain connectomes.}\label{adni}
\vspace{-2pt}
\end{figure}

\textbf{Group Analysis of Estimated Effective Connectivity.}
For each group, we also estimate the brain network with all the subject's data for 10 trials and average the learned $\mathbf{A}$ for producing the DAG with post-processing. The estimated effective connectivity networks are shown in Figure~\ref{adni}. For a better view of brain connectome, we set the threshold $\varepsilon$ to 0.015. In the Appendix, we also provide additional quantitative comparison results. Overall, we make several remarkable observations by comparing the two groups. (1) Compared with the CN group's effective connectivity, it is obvious that the number and strength of edges in AD patients decrease evidently. This observation is supported by a wide range of existing literature~\cite{article36,article39,article16}. (2) According to~\cite{article53}, the degeneration of basal ganglia (node 14) is considered to be a contributor to the cognitive decline in AD patients. We also identify this brain ROI as the most important biomarker in Alzheimer’s disease, as shown in Table~\ref{tab3}. Moreover, as illustrated in the 14th column of the adjacency matrices in Figure~\ref{adni}, the basal ROI of CN indeed demonstrates much more connections with other brain regions than that of AD. (3) Many previous works~\cite{article49,article50} have found degenerative changes and reduced functional activity of the anterior cingulate cortex (ACC) in AD. We also identify ventral ACC (node 24) as the second most salient biomarker and find that the ventral ACC of AD has fewer and weaker neural links with other ROIs than that of CN, as shown in the 24th column of the adjacency matrices in Figure~\ref{adni}. Due to the page limit, we provide some other findings in the Appendix. 

The change of effective connectivity is believed to be highly related to brain disease~\cite{article63,article36}. Our observations verified by existing neuropathological studies indicate that ST-DAGCN provides a promising perspective for identifying the biomarkers associated with Alzheimer’s disease, which is worth being further studied to understand the underlying mechanisms in the future.

\section{Conclusion}
In this paper, we propose the ST-DAGCN model to jointly infer the brain effective connectivity and classify rs-fMRI time series. In ST-DAGCN, DAG convolution and 1D convolution are interleaved for spatial-temporal nonlinear structural equation modeling. A score-based optimization problem is formulated for DAG-based effective connectivity learning and solved by the augmented Lagrangian method. We demonstrate that ST-DAGCN achieves improved gender classification performance against existing approaches on two public rs-fMRI datasets, and yields more interpretability in understanding brain activity patterns and pathology of brain disease such as Alzheimer’s disease.

\begin{ack}
This work was supported in part by the National Natural Science Foundation of China under Grant 62250055, Grant 62371288, Grant 62320106003, Grant 61932022, Grant 61971285, Grant 62120106007, and in part by the Program of Shanghai Science and Technology Innovation Project under Grant 20511100100. 
\end{ack}

\bibliographystyle{plain}
\bibliography{refs}

\newpage
\section*{Appendix}
\appendix

\section{Implementation Details}
Following ST-GCN~\cite{article8}, we train ST-DAGCN on the sub-sequences sampled from rs-fMRI time series. Each sub-sequence starts at a random time frame within its corresponding rs-fMRI times series. In the test procedure, the final gender classification for a target subject is determined by the voting (i.e., the average sigmoid values) of its $S$ randomly sampled sub-sequences' predictions. For the HCP dataset, we follow~\cite{article8} to set the length of sub-sequence $T^\prime$ to 128 and the number of voters $S$ to 64. For ADNI, we set $T^\prime=100$ and $S=64$ that empirically achieve good performance.

The other hyperparameter settings are empirically tuned to achieve the best performance and are summarized as follows. For the augmented Lagrangian, we use $\beta=10$ and $\gamma=1/4$, same as~\cite{article18}. The Adam optimizer with a learning rate of 0.001 and batch size of 64 is adopted to solve the subproblem~\eqref{eq4}. The balancing hyperparameter $\lambda$ for sparsity-based regularization is 0.001. The dropout rate is set to 0.5 and the weight decay of $L_2$ regularization is set to 0.001.

\section{Evaluation Metrics and Scheme}
Considering that gender prediction is a binary classification task, we adopt four common metrics to evaluate the prediction performance, i.e., accuracy (ACC), sensitivity (SEN), specificity (SPE), and the area under the receiver operating characteristic curve (AUC). To ensure the credibility of the classification results, 5-fold cross validation is implemented 5 times for each study and we report the mean and standard deviation of these evaluation metrics.

For effective connectivity learning, since there is no ground truth on the employed real-world rs-fMRI databases, we just provide qualitative analysis to verify the learned brain structures based on previous neuroscience research, similar to~\cite{article8,article35,article36,article57,article16,article62}. Moreover, following~\cite{article36,article16,article62}, we perform group analysis on the ADNI database by estimating the effective connectivity of the CN and AD groups, respectively, and comparing their differences. We also verify some of the important differences according to the existing literature on studying Alzheimer’s disease.

\section{Ablation study of Effective Connectivity}
\begin{figure}[ht]
\centerline{\includegraphics[scale=0.42]{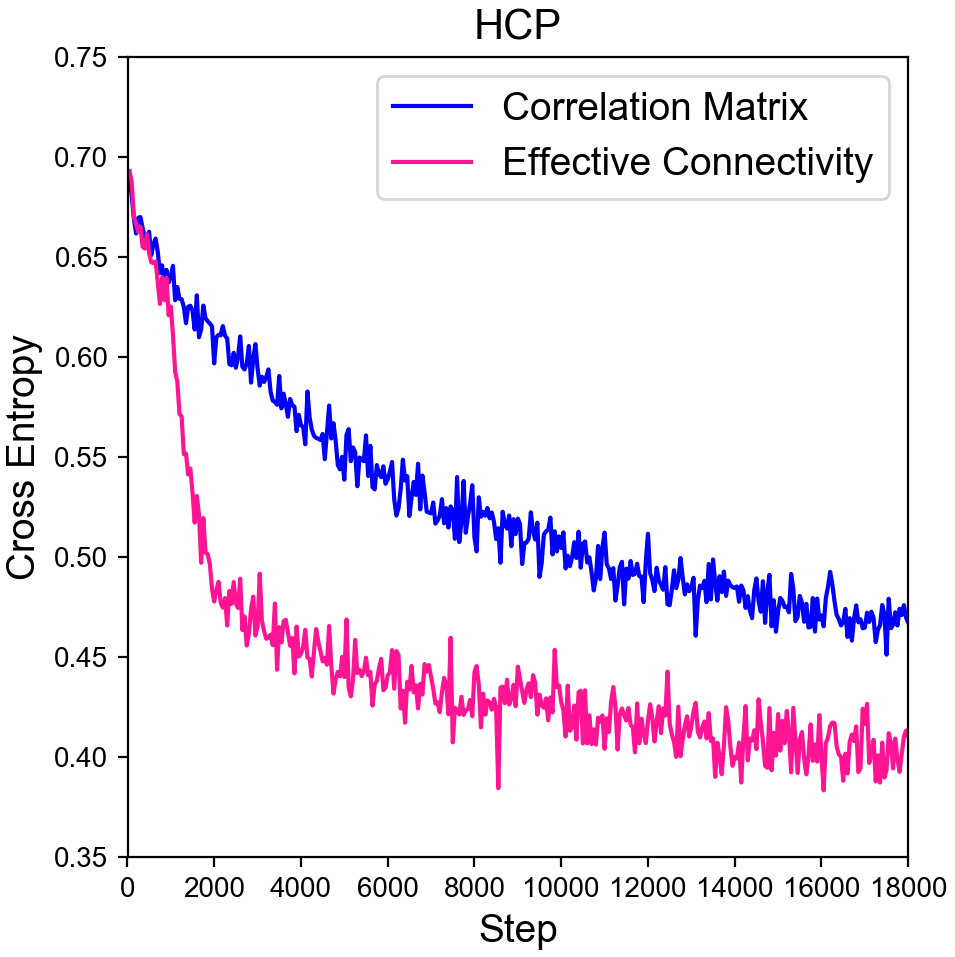}}
\caption{Cross entropy curves of ST-DAGCN in solving \eqref{eq4} and its counterpart model with brain connectivity set as the correlation matrix.}\label{loss}
\end{figure}

As our defined score function is based on the cross entropy between the model's predictions and the target labels, we provide the corresponding learning curves in Figure~\ref{loss} to verify that our causal discovery procedure indeed outputs a brain effective connectivity that benefits rs-fMRI time series classification. By solving~\eqref{eq4}, the cross entropy curves of ST-DAGCN drop much more quickly to smaller values than the counterpart ST-GCN model that uses the correlation matrix as brain connectivity. This result further demonstrates that a more meaningful brain connectivity modeling can lead to better rs-fMRI classification performance.

\section{Supplementary Results for Group Analysis on ADNI}
To quantify the difference between the two estimated effective connectivity networks, we calculate the difference at both the node level and edge level. Specifically, given the brain graph adjacency matrices of AD and CN groups $\mathbf{A}_{\text{AD}}$ and $\mathbf{A}_{\text{CN}}$, the differences of every node and every edge are:
\begin{align}
\text{Difference of node $v_i$} = \sum\nolimits_{j=1}^{N}\left|\mathbf{A}_{\text{AD}}[i,j]-\mathbf{A}_{\text{CN}}[i,j]\right| \nonumber\\
+ \sum\nolimits_{j=1}^{N}\left|\mathbf{A}_{\text{AD}}[j,i]-\mathbf{A}_{\text{CN}}[j,i]\right|, \label{node_diff}\\
\text{Difference of edge $v_i\rightarrow v_j$} = \left|\mathbf{A}_{\text{AD}}[i,j]-\mathbf{A}_{\text{CN}}[i,j]\right|.\label{edge_diff}
\end{align}

\begin{figure}[t]
\centerline{\includegraphics[scale=0.26]{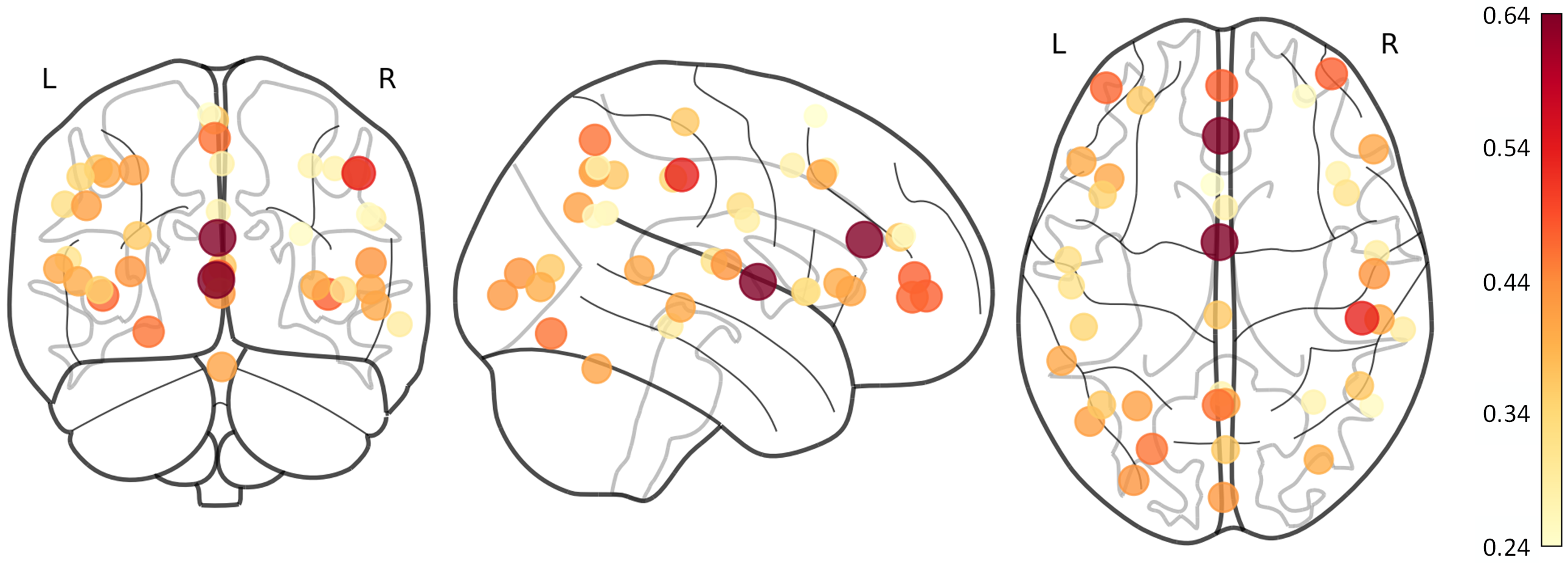}}
\caption{Node difference values by comparing AD versus CN.}\label{biomarker}
\end{figure}

\begin{figure}[t]
\centerline{\includegraphics[scale=0.24]{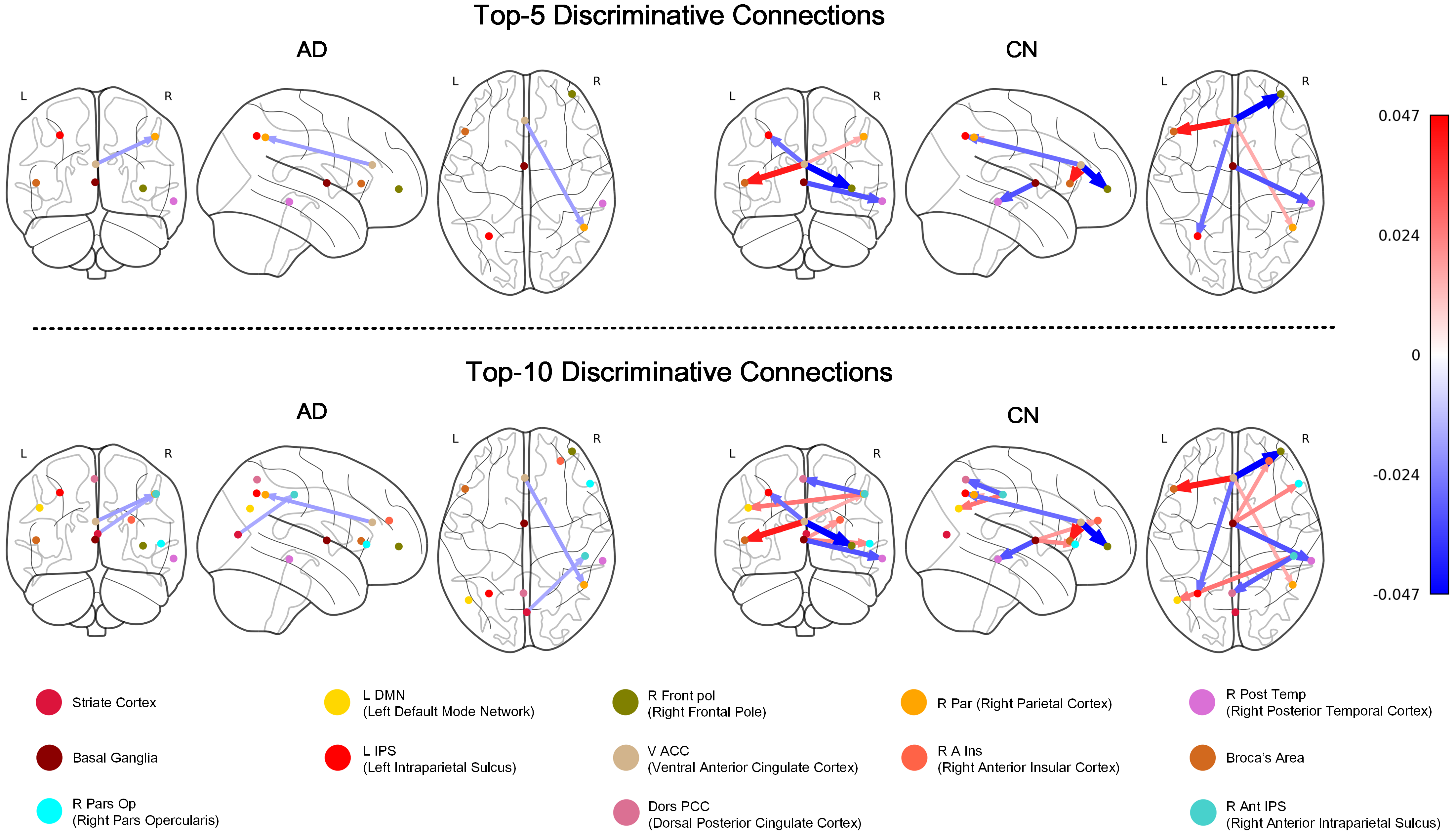}}
\caption{Discriminative connections by comparing AD versus CN.}\label{bioedge}
\end{figure}

We visualize the node difference values computed by Eq.~\eqref{node_diff} in Figure~\ref{biomarker} and list the top-10 salient brain ROIs in Table~\ref{tab3}. We name the selected ROIs as the regional biomarkers in identifying Alzheimer’s disease. The top-5 and top-10 discriminative brain graph edges are plotted in Figure~\ref{bioedge} based on Eq.~\eqref{edge_diff}, which are regarded as connectional biomarkers that differentiate between the AD and CN groups.
\begin{table}[t]
\setlength{\tabcolsep}{15pt}
\centering
\small
\caption{Top-10 salient brain ROIs in comparing AD versus CN.}\label{tab3}
\begin{tabular}{lcc}
\toprule
ROI Index & ROI Name  \\
\midrule
14 & Basal (basal ganglia) \\
24 & V ACC (ventral anterior cingulate cortex)  \\
39 & R Ant IPS (right anterior intraparietal sulcus)  \\
11 & R Front pol (right frontal pole)  \\
6  & Front DMN (front default mode network)  \\
17 & L Front pol (left frontal pole)  \\
34 & Dors PCC (dorsal posterior cingulate cortex) \\
21 & Vis (visual cortex)  \\
27 & R STS (right superior temporal sulcus)  \\
20 & L LOC (left lateral occipital cortex)  \\ 
\bottomrule
\end{tabular}
\end{table}

In addition to the biological discoveries illustrated in the main text, we have the following observations from group analysis: (1) We find from Figure~\ref{bioedge} that most discriminative connections for differentiating AD from CN involve ventral ACC, which also verifies that ACC is an important biomarker in identifying AD. (2) Besides Basal and ACC, some other identified biomarkers in Table~\ref{tab3} have also proved to relate to the differences between AD and CN. For instance, Reiner et al.~\cite{article60} found remarkable differences of cortical morphology in the IPS (node 39) between AD and CN. Finger et al.~\cite{article61} demonstrated that the reduced cortical thickness in the R Front pol (node 11) is associated with the presence of disinhibition in Alzheimer’s disease. Reduced metabolism in the PCC (node 34) and DMN (node 6) has been identified as a sign of Alzheimer's disease~\cite{article64,article32}. (3) Aligned with previous study~\cite{article51}, our estimated effective connectivity networks illustrate that the activation pattern of Broca's area (node 29) in AD group is weaker, such as $\text{ventral ACC}\rightarrow\text{Broca}$, which is inferred as the most discriminative connection with the largest edge difference value, as shown in Figure~\ref{bioedge}. (4) It is claimed that the metabolism in the temporoparietal junctions (TPJ, nodes 28 and 31) closely correlates with AD patients’ cognitive deficits~\cite{article52}. We find that the neural connections associated with TPJ also demonstrate some remarkable differences between AD and CN groups. For example, lesions in TPJ is demonstrated to relate to AD patients' impaired ability in visual perspective taking~\cite{article52}. In Figure~\ref{adni}, we find strong connection from visual region to the right TPJ (i.e., $\text{Vis}\rightarrow\text{R TPJ}$) in CN group while the corresponding link in AD group does not exist.

\section{Potential Future Work}
ST-DAGCN outputs a single effective connectivity for a group of subjects based on the optimization problem in~\eqref{eqn:opt} and ST-DAGCN is separately run for AD and CN respectively for group-level inference. In fact, existing works also separately infer the effective connectivity of AD and CN groups but treat male and female without distinction~\cite{article36,article16,article39}. We expect our paper can be further extended for jointly classifying Alzheimer’s disease and inferring two effective connectivity networks for the AD and CN groups. A potential method is to additionally learn two bias networks for the two groups, respectively. We can first learn a shared effective connectivity network across the two groups and then separately add the two bias networks to the shared network to obtain the effective connectivity of each group. The two effective connectivity networks are fed back to the model to further train the network to improve the rs-fMRI time series classification performance. The model can be iteratively refined using the recycling strategy~\cite{article59} that recursively computes the outputs and feeds the outputs into the model. We leave this work for future exploration.

\end{document}